\documentclass[runningheads]{llncs}
\usepackage[T1]{fontenc}
\usepackage{graphicx}
\usepackage{adjustbox}
\usepackage{tabularx}
\usepackage{booktabs}
\usepackage{placeins}
\usepackage{longtable}
\usepackage{multirow}
\usepackage{array}
\usepackage{float}
\usepackage{geometry}
\usepackage{amsfonts}
\usepackage{amssymb}
\usepackage{tabularx}
\usepackage{adjustbox}

\begin{document}
\title{RAPS-3D: Efficient interactive segmentation for 3D radiological imaging}
\author{Théo Danielou, Daniel Tordjman, Pierre Manceron, Corentin Dancette}
\institute{Raidium, France}

\maketitle             

\begin{abstract}
Promptable segmentation, introduced by the Segment Anything Model (SAM), is a promising approach for medical imaging, as it enables clinicians to guide and refine model predictions interactively.
However, SAM's architecture is designed for 2D images and does not extend naturally to 3D volumetric data such as CT or MRI scans. Adapting 2D models to 3D typically involves autoregressive strategies, where predictions are propagated slice by slice, resulting in increased inference complexity. Processing large 3D volumes also requires significant computational resources, often leading existing 3D methods to also adopt complex strategies like sliding-window inference to manage memory usage, at the cost of longer inference times and greater implementation complexity.
In this paper, we present a simplified 3D promptable segmentation method, inspired by SegVol, designed to reduce inference time and eliminate prompt management complexities associated with sliding windows while achieving state-of-the-art performance.

\end{abstract}

\section{Introduction}
The paradigm of promptable segmentation, exemplified by models like SAM~\cite{sam}, is increasingly influencing medical image analysis; however, effectively extending these capabilities to the complexities of 3D medical volumes presents ongoing challenges.

One approach to processing 3D data involves autoregressive methods, such as RadSAM~\cite{radsam} or MedSAM-2~\cite{medsam2}, where a 2D model iteratively propagates predictions across consecutive slices. Another approach consists in building a natively 3D model. For instance, SAM-Med3D~\cite{sammed3d} replaces the 2D encoder and decoder from SAM with 3D transformers and CNNs. SegVol~\cite{segvol} introduced a two-step \textit{zoom-out-zoom-in} method, demonstrating promising performance improvements. nnInteractive~\cite{nninteractive} adapts the semantic segmentation pipeline nnU-Net~\cite{nnUNet} for prompted segmentation with an efficient auto-zoom strategy.
However, a limitation of these existing approaches lies in the complex inference procedures. 
For instance, SegVol relies on sliding window techniques, necessitating several forward passes, while 2D autoregressive models require a pass for each slice. This results in prolonged inference times, hindering smooth interactive workflows, particularly during segmentation correction by radiologists.

To address these limitations, we develop a simple method inspired by SegVol's \textit{zoom-out-zoom-in} strategy and explicitly avoiding sliding window inference. Furthermore, we propose to use exclusively 2D bounding boxes, instead of 3D boxes used by other 3D models. This more intuitive prompt enhances the radiologist's experience during interactive segmentation. Our method aims to reduce inference overhead and improve the practicality of 3D medical image segmentation workflows.

\begin{figure}
    \centering
    \includegraphics[width=1\linewidth]{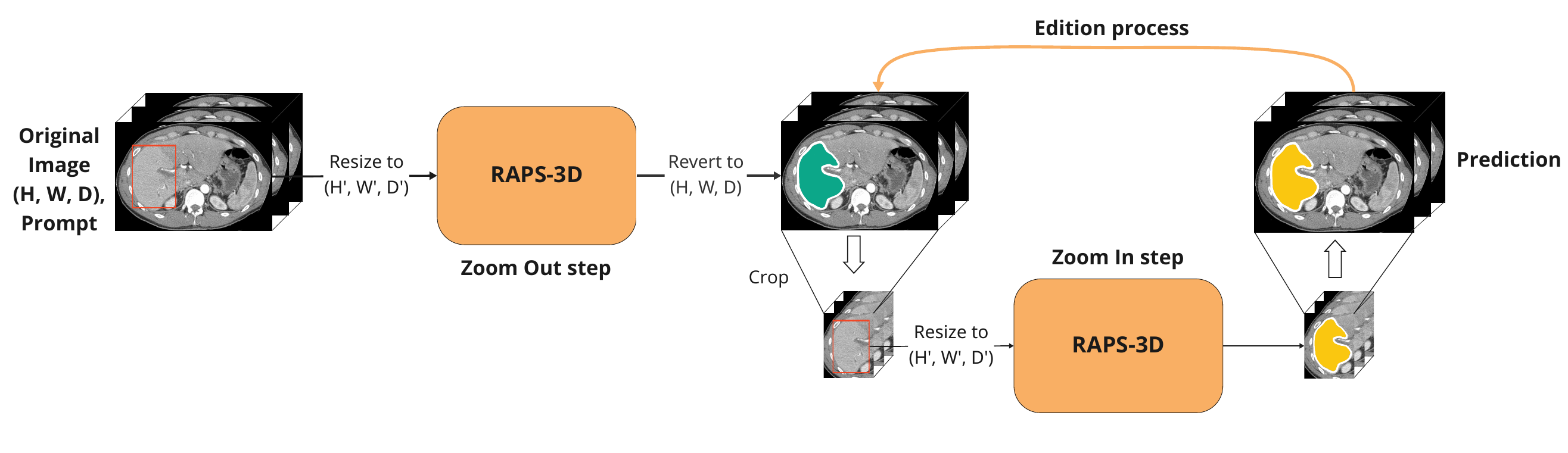}
    \caption{Overview of the \textbf{Raps-3D} inference method. (H, W, D) refers to the original image shape, and (H', W', D') refers to the input model shape.}
    \label{fig:enter-label}
\end{figure}

\section{Method}
We introduce Raps-3D (Radiological Prompted Segmentation-3D), a 3D segmentation model that takes as input a volumetric image of shape (H, W, D) and a prompt—either a point or a 2D bounding box—and outputs a 3D segmentation mask. \\
Raps-3D is a 3D adaptation of the SAM architecture, designed to process volumes of size (256, 256, 32) with patch sizes of (16, 16, 4). It features a prompt encoder that supports points, masks, and 2D bounding boxes. To improve usability and reduce the ambiguity often associated with 3D bounding boxes, we rely on 2D bounding boxes on a single slice as a simpler and more intuitive way for users to guide the model.

\subsection{Zoom-out-Zoom-in inference}
The model employs a simpler and more efficient \textit{zoom-out-zoom-in} strategy than proposed by SegVol. During \textit{zoom-out}, the full volumes are resized to the input size and then given to the model for a first step. \\
The main difference with the original \textit{zoom-out-zoom-in} relies on our \textit{zoom-in} step, which involves extracting the Region of Interest (ROI) defined by the initial zoom-out mask, cropping this area, resizing it to the model's input dimensions, and then feeding it back into the network for a second pass. In contrast, SegVol's method applied a sliding window with a fixed stride across this resized, cropped region. Our approach reduces the number of forward passes, increases the inference speed by design, facilitates the editing procedure, and mitigates prompt management issues while also promoting robustness to varying resolutions, consistent with the zoom-out training phase.

\subsection{Edition Procedure}
We implemented an efficient edition procedure. Upon radiologist identification of an error via a point prompt, the model refines the segmentation by processing this new prompt alongside the existing prompts and the current predicted mask. If the edit point occurs within the region of interest of the preceding zoom-in step, the already computed image features are reused to accelerate the re-inference process; otherwise, this region is expanded to encompass the new point plus a defined margin.

\subsection{Training details}
The training procedure incorporates both downsampled full volumes to (256, 256, 32), termed \textit{zoom-out} images, and cropped and resized regions to the ground truth, referred to as \textit{zoom-in} images. We observed that training also with \textit{zoom-out} images enhances inference performance, as a well-defined initial ROI significantly improves subsequent segmentation accuracy. Data augmentation for \textit{zoom-in} data includes variations in cropping margins and shifts. We use and fine-tune SegVol's pretrained VIT. CT data are normalized by clipping to [-500, 1000] Hounsfield Units and then applying \textit{MinMax} scaling to [0, 1]. The model was trained over 3 days on 8 nodes of the \textit{CINES Adastra HPC (High Performance Computing)}, each equipped with 4 AMD MI250X accelerators, for a total of 300 epochs. The learning rate was set to 1e-04, the batch size was equal to 8, and we used the Adam optimizer. \\

Our training dataset comprised 12,628 CT scans sourced from publicly available datasets: AbdomenAtlas \cite{abdomenatlas}, TotalSegmentator v1 \cite{totalsegmentator}, KiTS23 \cite{kits23}, MSD \cite{msd} (specifically Liver, Lung, Pancreas, and Colon), and the ULS23 challenge datasets \cite{uls23}. This collection includes 6,399 CT scans with annotations for organs and bones, and 6,299 CT scans featuring lesion annotations.

\section{Results and Discussion}

\begingroup
\setlength{\tabcolsep}{5pt}

\begin{table}[ht]
    \centering
    \caption{Performance (Mean Dice in percentage) on Validation set of AMOS-CT~\cite{amos} compared to 3D SAM's approaches. While SAM-Med3D doesn't report performance with 3 points, they achieve a Mean Dice of 83.99\% on AMOS-CT using 10-point prompts. The SegVol results were obtained using their publicly available pipeline demonstration and released model.}
    \begin{tabular}{llcc}
    \toprule
    Model & Prompt & Dice (Single prompt) & Dice (3 edition points) \\
    \midrule
    SAM-Med3D  & Point & 79.94 & --  \\
    \midrule
    \multirow{2}{*}{SegVol} & Point + Text & 66.49 & 69.77 \\
     & 3D Bbox + Text & 83.19 & --  \\
    \midrule
    \multirow{2}{*}{RadSAM}  & 3 Points & 84.06 & 86.39 \\
      & 3D Bbox & 85.02 & 87.32 \\
    \midrule
    \multirow{2}{*}{\textbf{Raps-3D}} & Point & 79.15 & 82.98 \\
     & 2D BBox &  \textbf{86.83} & \textbf{88.71} \\
    \bottomrule
    \end{tabular}
    \label{tab:my_label}
\end{table}
\endgroup

We evaluate Raps-3D on the validation set of the AMOS dataset, a widely used benchmark for 3D promptable organ segmentation. With a single 2D bounding box prompt, Raps-3D achieves state-of-the-art performance. It reaches a Dice score of 0.868 [0.862, 0.874], compared to 0.832 [0.824, 0.840] for SegVol using a 3D bounding box and text prompt (Wilcoxon signed-rank test, p < 0.001), and also outperforms RadSAM’s 3D bounding box-based approach. As for other models, our efficient editing strategy allows for 2-3\% increase in mean Dice score with only a few points. \\

In conclusion, we introduced a simple yet effective 3D promptable segmentation model that achieves performance on par with or better than current state-of-the-art methods. The model relies on an intuitive prompting and editing strategy, making it particularly well-suited for radiologists' clinical use.
It supports seamless interaction during editing while maintaining minimal complexity in prompt specification.
Future work will include evaluations on more datasets and additional imaging modalities such as MRI.

\section{Acknowledgments}
This project was provided with computing AI and storage resources by GENCI at CINES thanks to the grant 2024-AD011015473 on the supercomputer Adastra's MI250x partition.

\bibliographystyle{splncs04}

\bibliography{references}

\end{document}